\documentclass[sigconf, nonacm]{acmart}
\AtBeginDocument{%
  }

\renewcommand\footnotetextcopyrightpermission[1]{}
\usepackage{subcaption}
\usepackage{makecell}
\begin{document}

\title{PRISM: LLM-Guided Semantic Clustering for High-Precision Topics}

\author{Connor Douglas}
\affiliation{%
  \institution{New York University}
  \city{New York}
  \state{NY}
  \country{USA}}
\email{cpd8405@nyu.edu}

\author{Utkucan Balci}
\affiliation{%
  \institution{Binghamton University}
  \city{Binghamton}
  \state{NY}
  \country{USA}
}
\email{ubalci1@binghamton.edu}

\author{Joseph Aylett\textendash Bullock}
\affiliation{%
 \institution{United Nations}
  \city{New York}
  \state{NY}
  \country{USA}}
  \email{joseph.bullock@un.org}

\renewcommand{\shortauthors}{Connor Douglas, Utkucan Balci, \& Joseph Aylett–Bullock}

\begin{abstract}
   In this paper, we propose \textbf{Pr}ecision-\textbf{I}nformed \textbf{S}emantic \textbf{M}odeling \textbf{(PRISM)}, a structured topic modeling framework combining the benefits of rich representations captured by LLMs with the low cost and interpretability of latent semantic clustering methods. PRISM fine-tunes a sentence encoding model using a sparse set of LLM-provided labels on samples drawn from some corpus of interest. We segment this embedding space with thresholded clustering, yielding clusters that separate closely related topics within some narrow domain. Across multiple corpora, PRISM improves topic separability over state-of-the-art local topic models and even over clustering on large, frontier embedding models while requiring only a small number of LLM queries to train. This work contributes to several research streams by providing (i)~a student-teacher pipeline to distill sparse LLM supervision into a lightweight model for topic discovery;
     (ii)~an analysis of the efficacy of sampling strategies to improve local geometry for cluster separability; and 
    (iii)~an effective approach for web-scale text analysis, enabling researchers and practitioners to track nuanced claims and subtopics online with an interpretable, locally deployable framework.
\end{abstract}




\maketitle
\renewcommand{\thefootnote}{}
\footnotetext{To appear in \textit{Proceedings of the ACM Web Conference 2026 (WWW~'26)}, April 13--17, 2026, Dubai, United Arab Emirates.}
\renewcommand{\thefootnote}{\arabic{footnote}}

\section{Introduction}
Computationally identifying distinct narratives in social media posts is challenging.
This problem can be viewed as topic modeling with varying precision in topic definition (i.e., focus on macro topics, or more nuanced discourse).\footnote{With PRISM, topics are operationalized as clusters in a learned semantic space; keywords are not directly modeled but can be extracted \textit{post hoc} from clusters.}
While modern embedding-based topic modeling approaches~\cite{angelov2020top2vec,grootendorst2022bertopic} leverage rich semantic representations, they still often lack local precision for discerning subtle nuances within a given topical domain.
This is a critical problem when researchers or practitioners seek to track precise topics, in some cases involving sensitive areas of public security (e.g., extremist narratives or global crises). Recent work~\cite{zhangclusterllm,viswanathan2024large,pattnaik2024improving} has attempted to bridge this gap by using Large Language Models (LLMs) to guide the process of clustering text into topics. 
While these approaches improve topic quality by leveraging fine-grained representation captured by large models, reliance on LLMs in the final clustering pipeline introduces a dependency that hinders scalability, contributing to latency and high costs in the topic modeling process. To address this gap, we propose Precision-Informed Semantic Modeling (PRISM), a novel student-teacher framework that efficiently distills LLM-level semantic resolution into a lightweight encoder, achieving zero LLM cost at inference time via highly refined semantic vector-space separation and thresholded, fast clustering.\footnote{Python implementation available at: \href{https://github.com/connordouglas10/PRISM}{https://github.com/connordouglas10/PRISM}}
\\
\textbf{Related Work.}
Classical topic models like LDA~\cite{blei2003latent} represent documents as combinations of word-based topics, but their bag-of-words assumption and reliance on count statistics limit their ability to distinguish subtle sub-topics in short or domain-focused corpora. 
Embedding-based topic models like Top2Vec~\cite{angelov2020top2vec} and BERTopic~\cite{grootendorst2022bertopic} address these limitations by identifying dense clusters in a semantic vector space. 
A comparative study using Twitter data~\cite{egger2022topic} demonstrates that these methods yield more coherent and informative topics than classical topic models, such as LDA, when applied to social media corpora.
Prompt-based frameworks including TopicGPT~\cite{pham2024topicgpt} and PromptTopic~\cite{wang2023prompting} instead use LLMs at inference time to uncover topics. 
LLM guided clustering methods include ClusterLLM~\cite{zhangclusterllm}, which leverages triplet and pairwise queries to an instruction-tuned LLM to refine a small embedder and select an appropriate cut in a clustering hierarchy. ~\citet{viswanathan2024large} incorporate LLM feedback in the form of keyphrase expansion, pairwise constraints, and LLM post-correction to improve semi-supervised clustering. 
These approaches improve interpretability and alignment with user intent, but depend on repeated LLM calls during the clustering phase at inference.
Our work also connects to student-teacher architectures and the use of LLMs as labelers. 
~\citet{hoyle2020improving} provide empirical evidence that knowledge distillation can enhance neural topic model performance. Specific encoding models, like Sentence BERT~\cite{reimers2019sentence}, show that supervision on sentence similarity labels can yield embeddings well suited for clustering, while ranking-based loss functions like CoSENT~\cite{huang2024cosent} further refine sentence representations. 
Other streams have shown LLMs provide high-quality labels at lower cost than human annotation~\cite{pmlrv239mohta23a}.


\section{Model}

From a given corpus $C_\text{Text}$, we sample items and query a teacher model $L$ (large embedding or generative) to provide labels on item similarity.
These labeled samples comprise a dataset $D$, used to fine-tune a small, locally deployable student model $S$.
This fine-tuning process adapts the representation function $S$ to the particular domain of $C_{\text{Text}}$. 
Once $S$ has been tuned to the domain of $C_{\text{Text}}$, each item in $C_{\text{Text}}$ is encoded by $S$ to arrive at an embedded corpus $C_{\text{Emb}}$. Optionally, these embeddings are used with the labels generated by the teacher $L$ in $D$ to find an optimal threshold $\tau$ for clustering. A thresholded clustering algorithm $A$ then groups items in $C_{\text{Emb}}$.





\textbf{Model Training.}
PRISM fine-tunes a pre-trained model $S_0$ taken from the SentenceTransformers~\cite{reimers2019sentence} distribution in Python. 
We opt to start with the \texttt{all-mpnet-base-v2} model. This model is fine-tuned on a dataset generated by a teacher model $L$.

\textbf{Clustering.}
PRISM uses thresholded clustering techniques, which allow for unassignment of items, central to the goal of precise narrative discovery. Full partitioning of the embedding space into non-singleton topical regions risks collapsing nuances in item content. We use the \texttt{community detection} algorithm provided by SentenceTransformers, denoted as $A$, which clusters all items within $\tau$ similarity of some central point, where $\tau$ can be set manually or automatically by optimizing metrics on data labeled by $L$, as shown later.\footnote{This algorithm enjoys considerably better time complexity (worst case $\mathcal{O}(|C|^2)$) than the canonical hierarchical clustering ($\mathcal{O}(|C|^3)$).} 

\subsection{Fine-tuning Approaches}
The fine-tuning process is central to the PRISM architecture. We examine different approaches for generating samples from a large language model $L$, which would be costly to run continuously at inference time. The principal goal of fine-tuning is to expand the resolution of the representation by $S_0$ in the subspace described by $C_{\text{Text}}$. This way, we use a general-purpose embedding model and shift the parameters to better discern topics in the domain of interest. We tune $S_0$ into the tuned model $S$ on a dataset $D$ generated by the large model $L$. 
We describe sampling approaches to construct $D$, which are both used with CoSENT loss~\cite{huang2024cosent}.\footnote{We find CoSENT loss outperforms contrastive loss functions even for binary comparison datasets.}

\textbf{Binary Comparison.}
We query a generative LLM with a zero-shot prompt:\\
\texttt{"Answer in 'Yes' or 'No'. At a high level of detail, are these two pieces of text saying essentially the same thing: i and j ?"}.\\
Here, $i$ and $j$ are two items drawn from $C_{\text{Text}}$, where the completion produces a binary label $y$, equal to $1$ if the (post-processed) completion contains \texttt{"yes"} and $y=0$ otherwise. These samples are drawn from $C_{\text{Text}}$ without replacement; we refer to this dataset as \textbf{full-range (FR) comparison}. 
An alternative binary sampling technique is \textbf{range-bound (RB) comparison}, which draws a sample of pairs $i,j$ where $sim(S_0(i),S_0(j))$ is restricted to some interval where $y$ could not be trivially predicted by $S_0$. Empirically, with $S_0$ as \texttt{all-mpnet-base-v2}, this range is selected to be $[.65,.95]$. The resulting datasets are denoted $D_{\text{FR}}$, $D_{\text{RB}}$, respectively.

\textbf{Embedding Similarity.}
This method makes calls to a large embedding model, rather than to a generative model. In this approach, $n$ samples are drawn from $C_{\text{Text}}$ and embedded by the large model $L$. Then, pairwise similarity comparisons are generated to arrive at a dataset comprised of triples $(i,j,y)$, with $y = sim(L(i),L(j))$. These triples collectively comprise $D_{\text{Emb}}$, of the size $n(n-1)$.
\begin{figure*}[!t]
  \centering
  \begin{subfigure}{0.32\textwidth}
    \centering
    \includegraphics[width=.9\linewidth]{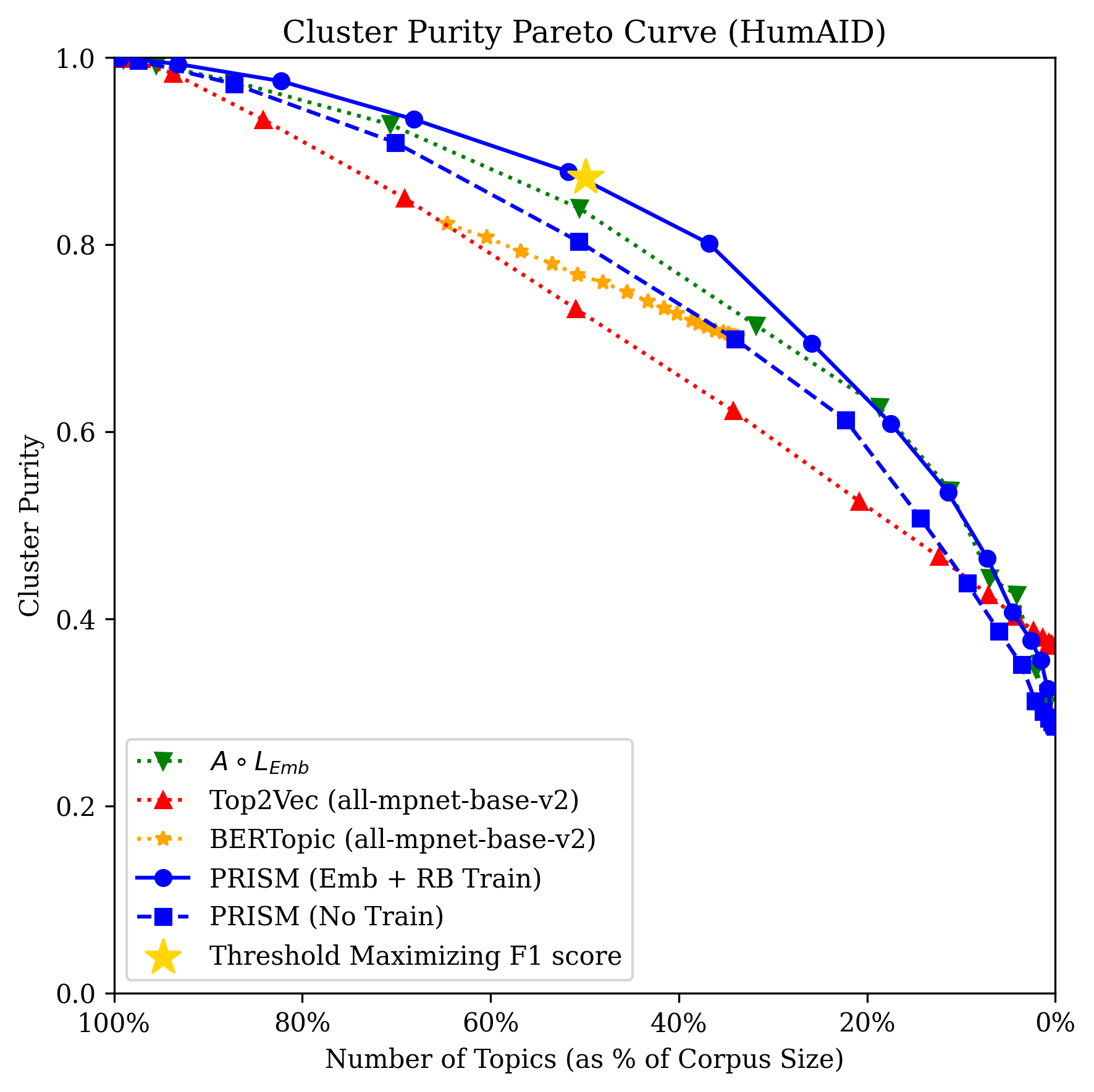}
    \caption{HumAID Corpus}
    \label{fig:humaid_purity_curve}
  \end{subfigure}
  \hfill
  \begin{subfigure}{0.32\textwidth}
    \centering
    \includegraphics[width=.9\linewidth]{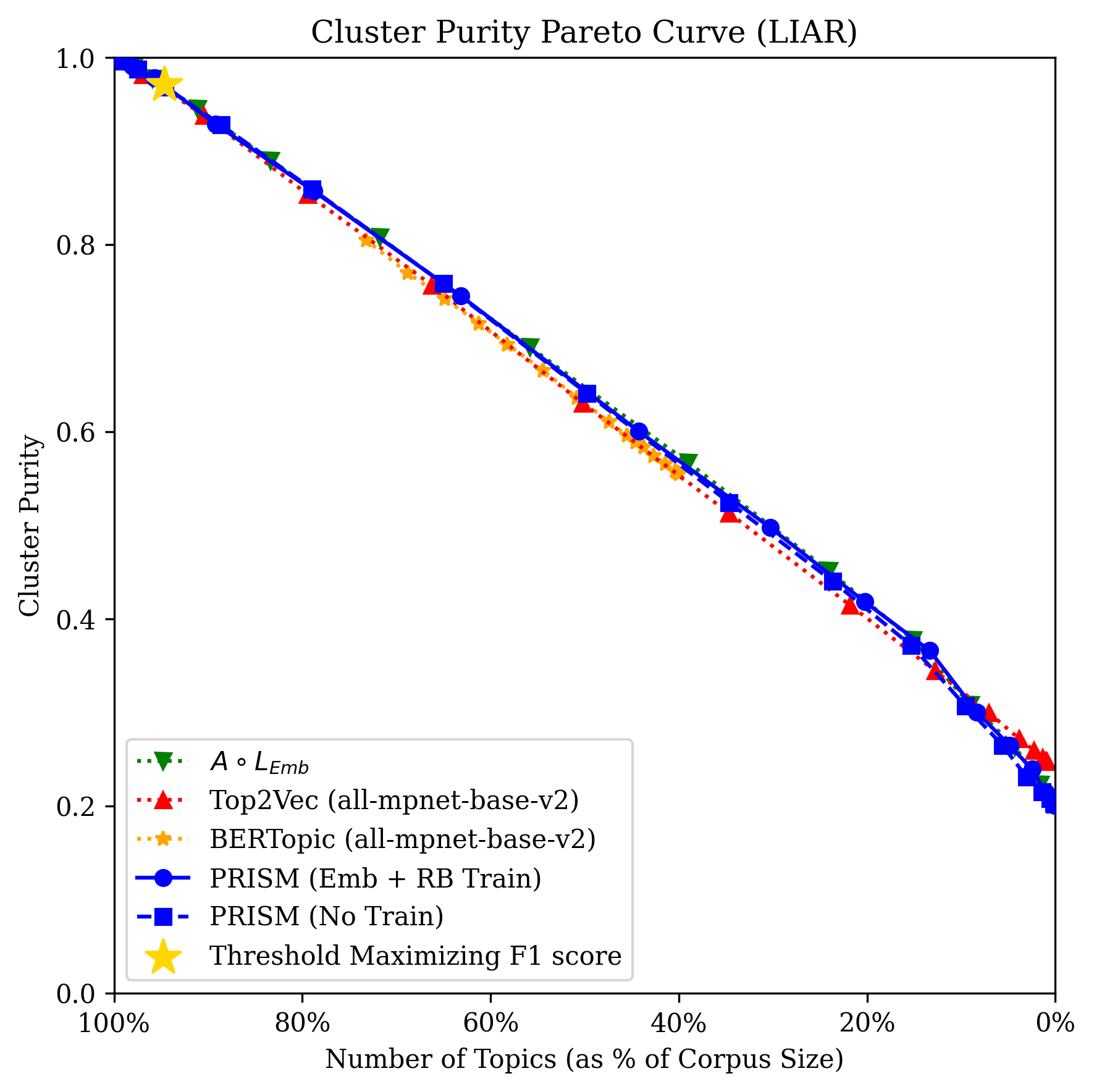}
    \caption{LIAR Corpus}
    \label{fig:liar_purity_curve}
  \end{subfigure}
  \hfill
  \begin{subfigure}{0.32\textwidth}
    \centering
    \includegraphics[width=.9\linewidth]{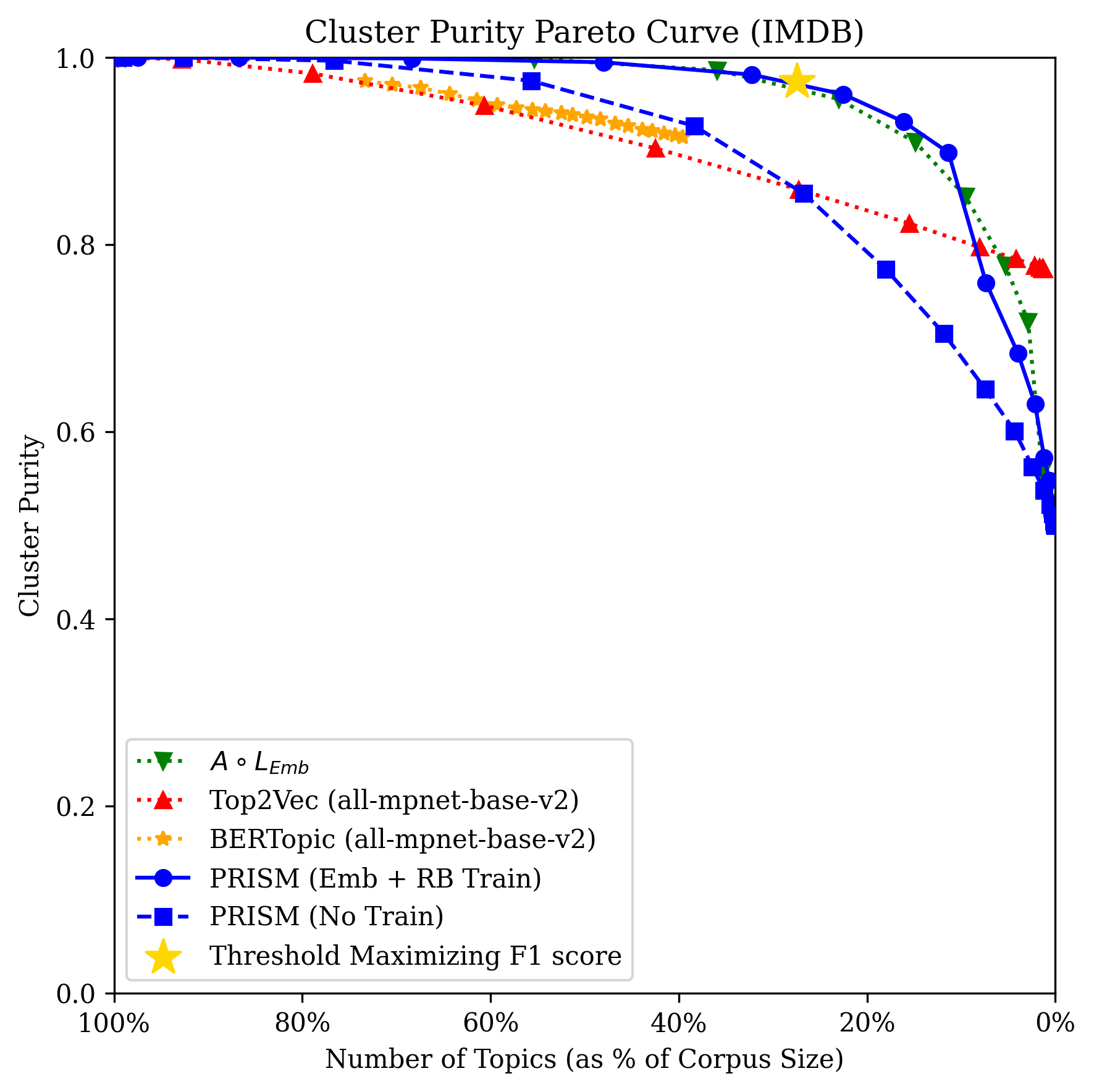}
    \caption{IMDB Corpus}
    \label{fig:imdb_purity_curve}
  \end{subfigure}
  \caption{Pareto curves depicting cluster purity against number of topics across three corpora.}
  \label{fig:pareto_curves}
\end{figure*}

\section{Experimental Results}
We now demonstrate the performance across various configurations and tuning procedures. To generate embedding datasets, OpenAI's \texttt{text-embedding-3-large} is used, denoted as $L_{\text{Emb}}$; for comparisons, $L_{\text{Comp}}$ is defined as OpenAI's \texttt{gpt-5-2025-08-07}.

\subsection{Experimental Setup}
\begin{figure}[t]
    \centering
    \includegraphics[width=.9\linewidth]{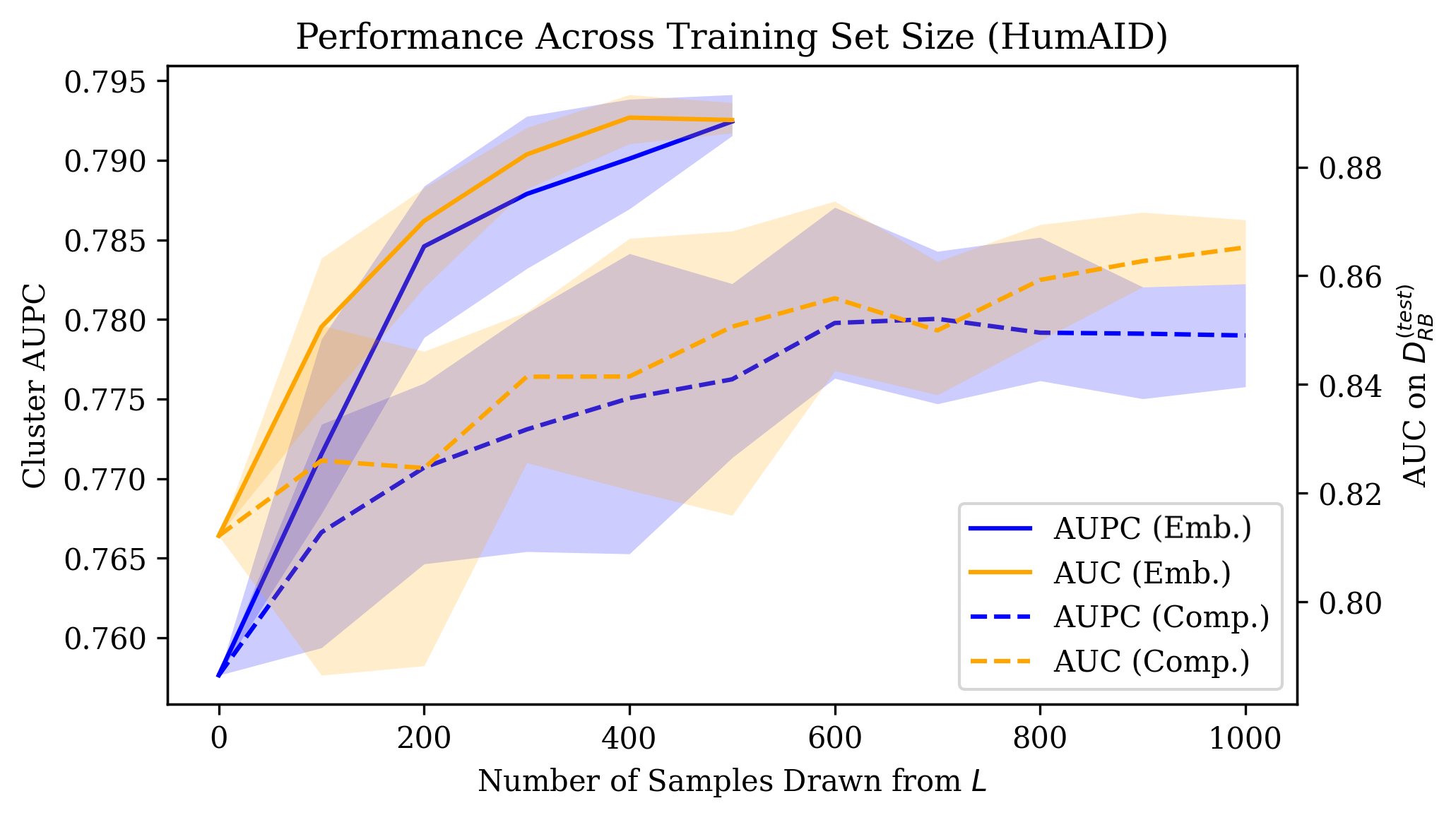}
    \caption{AUC and AUPC across training data size on HumAID for comparison and embedding dataset.}
    \label{fig:humaid_train_size}
\end{figure}

We compare PRISM against the two leading semantic clustering algorithms, \textbf{Top2Vec} and \textbf{BERTopic}, using \texttt{all-mpnet-base-v2} as the embedding model, which improves performance over each method's faster default model, to ensure a fair comparison. We also compare using the thresholded clustering algorithm directly on embeddings provided by the large teacher embedding model $L_{\text{Emb}}$ (\textbf{$A \circ L_{\text{Emb}}$}). This comparison isolates the benefit of PRISM's distillation techniques by benchmarking against direct clustering in the high-capacity embedding space of one teacher model, which avoids generative inference-time calls, but requires costly embeddings from a black-box model for each item. We fit PRISM and examine various performance metrics across three textual corpora. \textbf{HumAID (HU)} consists of over 77K disaster-related human-labeled tweets from 19 events between 2016 and 2019~\cite{alam2021humaid}. 
\textbf{LIAR (LI)} is a fact-checking dataset~\cite{wang2017liar} consisting of 12.8K short political statements gathered from politifact.com, labeled for truthfulness (with 6 degrees of truthfulness). 
\textbf{IMDB (IM)} consists of 50K movie reviews with balanced positive and negative labels~\cite{maas2011learning}, allowing us to test PRISM's performance on discerning opposing opinions as distinctive topics.~\footnote{After post-processing, corpora (train, test) sizes are: HU$=(28{,}802, 8{,}157)$,LI$=(10{,}240, 2{,}551)$, and IM$=(25{,}000, 10{,}000)$, where we constrain the test size of IM from $25,000$ with a balanced sample.} Training sets are generated for each corpus, yielding $|D_{\text{RB}}|=|D_{\text{FR}}| = 1,000$ and $|D_{\text{Emb}}|=249,500$, created from $500$ embeddings.
~\footnote{All datasets are trained using CoSENT loss with a learning rate of $3 \times 10^{-6}$ over $5$ epochs. The cost of generating $D_{\text{RB}}$ and $D_{\text{FR}}$ is approximately \$0.20 each; $D_{\text{Emb}}$ is approximately \$0.05.} Analysis of training set size is presented in Figure~\ref{fig:humaid_train_size}, where we set final training sizes to be values where performance appears to flatten.~\footnote{We select sample sizes based on AUC performance on $D_{\text{RB}}^{(\text{val})}$, which is observable \textit{a priori} in the topic modeling process, rather than AUPC metric which would not be possible to calculate in unlabeled sets.}
Our work is not about explicitly labeling data; the labels of these datasets are therefore only used as a ground-truth proxy for practically relevant topics.


\subsection{Knowledge Distillation}
We first describe the experimental results of training $S_0$ using various approaches across $D_{\text{RB}}$, $D_{\text{FR}}$, and $D_{\text{Emb}}$, drawn from a training set of $C_{\text{Text}}$ and separate from the test set used for evaluation. These results provide a credible lower bound on PRISM's performance, where in practice the model may cluster samples that were also used for fine-tuning. To demonstrate that fine-tuning effectively distills knowledge from $L$ to $S$, for each corpus, we generate a range-bound dataset of comparisons $D_{\text{RB}}^{(\text{test})}$ from the holdout-test set ($n=1,000$), labeled by $L_{\text{Comp}}$. Treating similarities in the space defined by $S$ between pairs in $D_{\text{RB}}^{(\text{test})}$ as a class scoring model, we examine the area under the ROC curve (AUC). Results on this set are presented in Table \ref{tab:auc_results}. We see that fine-tuning using any approach substantially increases the model's discriminating power on labels generated by $L_{\text{Comp}}$ in $D_{\text{RB}}^{(\text{test})}$, demonstrating that a higher resolution representation can be effectively transferred from $L$ to $S$. We note that fine-tuning achieves comparable or superior performance to the large model $L_{\text{Emb}}$ directly. This may be due to $D_{\text{Emb}}$ querying $L$ fewer times, despite more resultant training samples, and to the training comparison datasets being labeled from the same model $L_{\text{Comp}}$ as $D_{\text{RB}}^{(\text{test})}$, whereas $D_{\text{Emb}}$ is labeled via $L_{\text{Emb}}$, a different, albeit related, model. To examine whether $S$ trained on comparison labels may overfit to $L_{\text{Comp}}$, we now look at downstream topic clustering performance on ground truth labels. As training on $D_{\text{RB}}$ yields the strongest performance advantage, we perform comparison fine-tuning for topic analysis only with $D_{\text{RB}}$ (not $D_{\text{FR}}$)  and construct another dataset $D_{\text{RB}}^{(\text{val})}$ as a validation set for threshold tuning.

\subsection{Topic Analysis}
To evaluate topic quality, we look beyond word-count metrics like topic coherence and diversity, which can overfit and fail to capture full semantic relations in $C_{\text{Text}}$, especially in confined domains with near-identical vocabularies.
Instead, we assume corpus labels reflect coarse-grained informational similarity: items sharing the same label (e.g., ‘Requesting Aid’ vs ‘Infrastructure Damage’ in HumAID, truthfulness score in LIAR, sentiment in IMDB) convey similar information. 
These labels provide a meaningful measure of informational similarity outside of word-based metrics. Well-formed topic clusters should avoid grouping items from conflicting labels. Using these labels, we examine the standard \textit{cluster purity} metric on resultant clusters.\footnote{Cluster purity treats each cluster as a classifier by majority vote and counts the portion of items correctly classified.} 
We use these labels \textit{only for evaluation} and measure cluster purity over items in $C_{\text{Text}}$.

\begin{table}
    \centering
\begin{tabular}{l|ccc|ccc}
\toprule
\makecell{}& \multicolumn{3}{c}{\textbf{mpnet + Training Approach}}& \multicolumn{3}{c}{\textbf{Model}} \\
\makecell{\textbf{$C_{\text{Text}}$}}& \makecell{FR} 
& \makecell{RB } 
& \makecell{Emb} 
& \makecell{mpnet} 
& \makecell{Mini} 
& \makecell{$L_{\text{Emb}}$} 
 \\
\midrule
HU & .806& .838 & \textbf{.892}& .812 & .789 & .891 \\
LI   & .745& \textbf{.824} & .792& .745 & .750 & .818\\
IM   & .906 & \textbf{.912} & .811&  .584 & .605 & .789 \\
\bottomrule
\end{tabular}
    \caption{AUC of models predicting labels in $D_\text{RB}^{(\text{test})}$; \textit{mpnet} corresponds to \textit{all-mpnet-base-v2}, \textit{Mini} corresponds to \textit{all-MiniLM-L12-v2}.}
    \label{tab:auc_results}
\end{table}

\textbf{Topic Precision.}
As clustering threshold $\tau$ increases, items must be more similar to merge, leaving slightly more items unassigned as singleton clusters.
For \textit{Top2Vec} and \textit{BERTopic}, each item is assigned a probability of belonging to the assigned cluster. By scanning across clustering/probability thresholds, we trace out cluster purity at varying numbers of clusters. Specifically, we analyze model performance using a \textit{Pareto curve},  which accounts for the confounding effect of cluster size, where many small clusters could create more accurate clusters (higher purity). This curve is directly motivated by an inherent multi-objective tradeoff: more clusters require fewer manual checks per cluster, but at the cost of coarser, more inaccurate clusters. Figure~\ref{fig:pareto_curves} shows the total number of clusters (including unassigned singletons) as a fraction of the total size of the corpus on the $x$-axis; the $y$-axis shows cluster purity.~\footnote{Note, the HDBSCAN in BERTopic will leave outlier items unclustered, so the curve will not extend to the edges.} 
 Here, we set $\tau$ to maximize F1 score on the holdout $D_{\text{RB}}^{(\text{val})}$ dataset (generated via $L_{\text{Comp}}$) to demonstrate automated threshold selection.

Across all corpora, we see that PRISM, when fine-tuned on both $D_{\text{RB}}$ and $D_{\text{Emb}}$, sits on or near the frontier nearly everywhere. PRISM strictly dominates BERTopic for any given number of clusters, while Top2Vec tends to perform slightly better only when the number of clusters is large. To aggregate performance, we denote a model's \textit{precision} to be its ability to produce highly pure clusters across all clustering thresholds, measured by the \textbf{area under the Pareto curve (AUPC)}. We display the full Pareto curve only for PRISM trained on both $D_{\text{Emb}}$ and $D_{\text{RB}}$ for visual clarity. However, we report the AUPC values for curves under various fine-tuning types. Since BERTopic does not span the $x$-domain and is strictly dominated by PRISM, we only compare AUPC of PRISM against Top2Vec and $A \circ L_{\text{Emb}}$. Table~\ref{tab:aupc_results} shows AUPC comparisons across topic models trained with different query approaches.
Across corpora, PRISM with domain-adapted tuning consistently achieves higher AUPC than Top2Vec, and even outperforming clustering with $L_{\text{Emb}}$ directly. 
In general, fine-tuning on both embedding ($D_{\text{Emb}}$) and comparison ($D_{\text{RB}}$) datasets tends to yield the most precise topic model. This improvement likely comes from both an increased training size and orthogonality in signal, which is generated from two distinct models $L_{\text{Emb}}$ and $L_{\text{Comp}}$. Importantly, we note that knowledge distillation from fine-tuning is central to the outperformance of our model, where untuned PRISM yields, at best, marginal gains over Top2Vec. This result verifies that performance improvements are not coming strictly from the clustering algorithm, but rather the joint PRISM modeling process.


\section{Conclusion}
In this work, we introduce PRISM, a semantic modeling framework that distills LLM representations into a discriminating, lightweight encoder for use in precision topic modeling and discovery. By combining LLM-supervised fine-tuning with thresholded clustering allowing for unassignment, PRISM avoids low quality assignments and effectively separates closely related narratives within some narrow topical domain. Across three diverse corpora, PRISM learns to distill qualitative representations from a teacher LLM into its embedding space, considerably improving AUC in predicting teacher similarity evaluations. The resulting topics on this refined embedding space achieve high cluster purity, effectively capturing nuanced informational characteristics reflected in human labels and improving on demonstrably strong baselines. These results suggest that small encoder models can effectively refine their embedding space to drastically improve in-domain semantic resolution and leverage this precision for deriving more semantically precise topical clusters. Further, PRISM's outperformance against directly clustering on $L_{\text{Emb}}$ indicates that careful querying of multiple models and subsequent fine-tuning can produce a state-of-the-art representation within domain, while being considerably cheaper at inference.

\begin{table}[t]
    \centering
    \setlength{\tabcolsep}{3pt}      
    \renewcommand{\arraystretch}{0.9} 
    \begin{tabular}{l|cccc|ccc}
    \toprule
     

    \makecell{}& \multicolumn{4}{c}{\textbf{PRISM + Training Approach}}& \multicolumn{3}{c}{\textbf{Topic Model}} \\
    \makecell{\textbf{$C_{\text{Text}}$}}
    & \makecell{No train} 
    & \makecell{RB} 
    & \makecell{Emb} 
    & \makecell{Emb.\\ \& RB} 
    & \makecell{T2V\\(mpnet)}
    & \makecell{T2V\\(d2v)}
    & \makecell{$A \circ L_{\text{Emb}}$}\\
    \midrule
    HU & .758& .782 & .792& \textbf{.807}&  .717& .679 & .786\\
    LI   & .635 & \textbf{ .641} & .636& .638& .636& .631 & .639 \\
    IM   & .900 & .948 & . 948 & \textbf{.966}& .913 & .886 & .957\\
    \bottomrule
    \end{tabular}
    \caption{AUPC results on labeled holdout test set (higher is better). Top2Vec uses \texttt{Doc2Vec} (d2v) as the default model; this is tested against Top2Vec using \texttt{all-mpnet-base-v2} (mpnet).}
    \label{tab:aupc_results}
\end{table}

\textbf{Future Work.}
A promising extension to PRISM is in active learning in the knowledge distillation process, enabled by low query-to-label latency. 
Another dimension worth exploring is extracting richer signals from LLMs, including learning representations from explanations and internal activations in open-weight models. Finally, we hope to see PRISM applied to enable richer domain-level findings in the social and computational sciences.
\bibliographystyle{ACM-Reference-Format}
\bibliography{references}

@article{blei2003latent,
  title={Latent Dirichlet Allocation},
  author={Blei, David M and Ng, Andrew Y and Jordan, Michael I},
  journal={Journal of Machine Learning Research},
  volume={3},
  number={Jan},
  pages={993--1022},
  year={2003}
}

@article{egger2022topic,
  title={A Topic Modeling Comparison Between LDA, NMF, Top2Vec, and BERTopic to Demystify Twitter Posts},
  author={Egger, Roman and Yu, Joanne},
  journal={Frontiers in Sociology},
  volume={7},
  pages={886498},
  year={2022},
  publisher={Frontiers Media SA}
}

@article{angelov2020top2vec,
  title   = {Top2Vec: Distributed Representations of Topics},
  author  = {Angelov, Dimo},
  journal = {arXiv:2008.09470},
  year    = {2020}
}

@article{grootendorst2022bertopic,
  title   = {{BERTopic}: Neural Topic Modeling with a Class-based {TF}-{IDF} Procedure},
  author  = {Grootendorst, Maarten},
  journal = {arXiv:2203.05794},
  year    = {2022}
}

@inproceedings{pham2024topicgpt,
  title={TopicGPT: A Prompt-based Topic Modeling Framework},
  author={Pham, Chau Minh and Hoyle, Alexander and Sun, Simeng and Resnik, Philip and Iyyer, Mohit},
  booktitle={NAACL},
  pages={2956--2984},
  year={2024}
}

@inproceedings{zhangclusterllm,
    title = "{C}luster{LLM}: Large Language Models as a Guide for Text Clustering",
    author = "Zhang, Yuwei  and
      Wang, Zihan  and
      Shang, Jingbo",
    
    booktitle = "EMNLP",
    month = dec,
    year = "2023",
    
    publisher = "Association for Computational Linguistics",
    
    pages = "13903--13920",
    
}

@article{viswanathan2024large,
  title={Large Language Models Enable Few-shot Clustering},
  author={Viswanathan, Vijay and Gashteovski, Kiril and Lawrence, Carolin and Wu, Tongshuang and Neubig, Graham},
  journal={TACL},
  volume={12},
  pages={321--333},
  year={2024},
  publisher={MIT Press }
}

@inproceedings{pattnaik2024improving,
  title={Improving Hierarchical Text Clustering with LLM-guided Multi-view Cluster Representation},
  author={Pattnaik, Anup and George, Cijo and Tripathi, Rishabh Kumar and Vutla, Sasanka and Vepa, Jithendra},
  booktitle={EMNLP},
  pages={719--727},
  year={2024}
}

@inproceedings{reimers2019sentence,
  title={Sentence-BERT: Sentence Embeddings using Siamese BERT-Networks},
  author={Reimers, Nils and Gurevych, Iryna},
  booktitle={EMNLP-IJCNLP},
  pages={3982--3992},
  year={2019}
}

@article{huang2024cosent,
  title={CoSENT: Consistent Sentence Embedding via Similarity Ranking},
  author={Huang, Xiang and Peng, Hao and Zou, Dongcheng and Liu, Zhiwei and Li, Jianxin and Liu, Kay and Wu, Jia and Su, Jianlin and Yu, Philip S},
  journal={IEEE/ACM Transactions on Audio, Speech, and Language Processing},
  volume={32},
  pages={2800--2813},
  year={2024},
  publisher={IEEE}
}

@InProceedings{pmlrv239mohta23a,
  title = 	 {Are Large Language Models Good Annotators?},
  author =       {Mohta, Jay and Ak, Kenan and Xu, Yan and Shen, Mingwei},
  booktitle = 	 { NeurIPS 2023 Workshops},
  pages = 	 {38--48},
  year = 	 {2023},

  volume = 	 {239},
  series = 	 {Proceedings of Machine Learning Research},
  month = 	 {16 Dec},
  publisher =    {PMLR},
}

@inproceedings{wang2023prompting,
  title={Prompting Large Language Models for Topic Modeling},
  author={Wang, Han and Prakash, Nirmalendu and Hoang, Nguyen Khoi and Hee, Ming Shan and Naseem, Usman and Lee, Roy Ka-Wei},
  booktitle={IEEE BigData},
  pages={1236--1241},
  year={2023},
  organization={IEEE}
}

@inproceedings{hoyle2020improving,
  title={Improving Neural Topic Models Using Knowledge Distillation},
  author={Hoyle, Alexander Miserlis and Goel, Pranav and Resnik, Philip},
  booktitle={EMNLP},
  pages={1752--1771},
  year={2020}
}

@inproceedings{wang2017liar,
    title = "``Liar, Liar Pants on Fire'': A New Benchmark Dataset for Fake News Detection",
    author = "Wang, William Yang",
    
    booktitle = "ACL",
    month = jul,
    year = "2017",
    address = "Vancouver, Canada",
    publisher = "ACL",
    pages = "422--426",
    
}

@inproceedings{alam2021humaid,
  title={Humaid: Human-annotated Disaster Incidents Data from Twitter with Deep Learning Benchmarks},
  author={Alam, Firoj and Qazi, Umair and Imran, Muhammad and Ofli, Ferda},
  booktitle={ICWSM},
  volume={15},
  pages={933--942},
  year={2021}
}

@inproceedings{maas2011learning,
  title={Learning Word Vectors for Sentiment Analysis},
  author={Maas, Andrew and Daly, Raymond E and Pham, Peter T and Huang, Dan and Ng, Andrew Y and Potts, Christopher},
  booktitle={ACL},
  pages={142--150},
  year={2011}
}

\end{document}